\documentclass[conference, a4paper]{IEEEtran}
\IEEEoverridecommandlockouts

\ifCLASSINFOpdf
  \usepackage[pdftex]{graphicx}
  \DeclareGraphicsExtensions{.pdf,.jpeg,.png}
\else
\fi

\usepackage{cite}
\usepackage{amsmath,amssymb,amsfonts}
\usepackage{algorithmic}
\usepackage{graphicx}
\usepackage{textcomp}
\usepackage[left=1.4cm, right=1.4cm, top=1.7cm]{geometry}
\usepackage{caption}
\usepackage{anyfontsize}
\usepackage{hyperref}
\usepackage{xcolor}

\makeatletter
\newcommand*\titleheader[1]{\gdef\@titleheader{#1}}
\AtBeginDocument{%
\let\st@red@title\@title
\def\@title{%
\bgroup\normalfont\normalsize\centering\@titleheader\par\egroup
\vskip0.2em\st@red@title}
}
\makeatother

\makeatletter
\renewcommand{\fnum@figure}{Figure \thefigure}
\makeatother

\title{{Learning to Explain Air Traffic Situation}  
\thanks{This work is supported by the Korea Agency for Infrastructure Technology Advancement (KAIA) grant funded by the Ministry of Land, Infrastructure and Transport (Grant RS-2022-00156364).}
}

\titleheader{First US-Europe Air Transportation Research and Development Symposium (ATRDS2025)}

\author{\IEEEauthorblockN{Hong-ah Chai, Seokbin Yoon, and Keumjin Lee}
\IEEEauthorblockA{Department of Air Transport, Transportation and Logistics\\
Korea Aerospace University\\
Goyang, South Korea \\
honga1534@naver.com, \{sierra.bin, keumjin.lee\}@kau.ac.kr}
}

\renewcommand\IEEEkeywordsname{Keywords}
\IEEEaftertitletext{\vspace{-1\baselineskip}}

\begin{document}

\maketitle

\noindent \begin{abstract}
Understanding how air traffic controllers construct a mental ‘picture’ of complex air traffic situations is crucial but remains a challenge due to the inherently intricate, high-dimensional interactions between aircraft, pilots, and controllers. Previous work on modeling the strategies of air traffic controllers and their mental image of traffic situations often centers on specific air traffic control tasks or pairwise interactions between aircraft, neglecting to capture the comprehensive dynamics of an air traffic situation. To address this issue, we propose a machine learning-based framework for explaining air traffic situations. Specifically, we employ a Transformer-based multi-agent trajectory model that encapsulates both the spatio-temporal movement of aircraft and social interaction between them. By deriving attention scores from the model, we can quantify the influence of individual aircraft on overall traffic dynamics. This provides explainable insights into how air traffic controllers perceive and understand the traffic situation. Trained on real-world air traffic surveillance data collected from the terminal airspace around Incheon International Airport in South Korea, our framework effectively explicates air traffic situations. This could potentially support and enhance the decision-making and situational awareness of air traffic controllers.
\end{abstract}

\vspace{0.3cm}

\begin{IEEEkeywords}
air traffic situation explanation; air traffic control; machine learning; multi-agent trajectory model
\end{IEEEkeywords}

\section{Introduction}
Air traffic situations involve the interdependence of multiple flights and the coordination between pilots and Air Traffic Controllers (ATCo), making them challenging to model and predict~\cite{porretta2008performance,bilimoria2005analysis}. To manage these complexities, ATCo develop a comprehensive and abstract mental representation of the air traffic situation, often referred to as a ‘picture’~\cite{nunes2003identifying}. This mental representation allows them to focus on the most important and relevant aspects of the air traffic situation while filtering out less critical information~\cite{histon2002introducing}.  

Various approaches have been proposed for designing automated Air Traffic Control (ATC) systems~\cite{kuchar2000review,beasley2000scheduling,pallottino2002conflict}. However, the results of these approaches often contradict how ATCo actually perform ATC tasks. As highlighted in~\cite{guleria2024towards}, solutions designed to mimic the decision-making patterns of human controllers, rather than solely relying on optimized solutions, tend to achieve greater operational efficiency and acceptance. Aligning with this perspective, recent works are increasingly focusing on developing data-driven approaches capable of learning ATCo’s actions from historical air traffic data~\cite{hong2015modeling,jung2018data}. 

While previous studies have successfully modeled ATC strategies, these efforts have predominantly concentrated on specific ATC tasks, such as conflict resolution or landing order sequencing, often limiting themselves to pairwise interactions between aircraft. However, managing air traffic situations is a more abstract and extensive problem that individual ATC action models and their combinations cannot effectively capture. In this regard, we propose a novel framework that explains air traffic situations using a machine learning technique. This paper, as the initial phase of our study, primarily focuses on discerning why particular air traffic flows become established and how aircraft within these flows impact one another. While traditional approaches are inadequate in holistically explaining air traffic situations, our model utilizes historical air traffic data to identify interdependencies between aircraft and measure their impact on overall traffic dynamics.

The remainder of this paper is organized as follows: Section~\ref{framework} introduces the proposed framework, detailing its approach and learning paradigm. Section~\ref{preliminary} presents preliminary results, explaining dynamic air traffic situations. Finally, Sections~\ref{discussion} and ~\ref{conclusion} conclude the paper and outline potential directions for future research.

\begin{figure*}[t!]			
	\centering
	\includegraphics[width=0.95\textwidth]{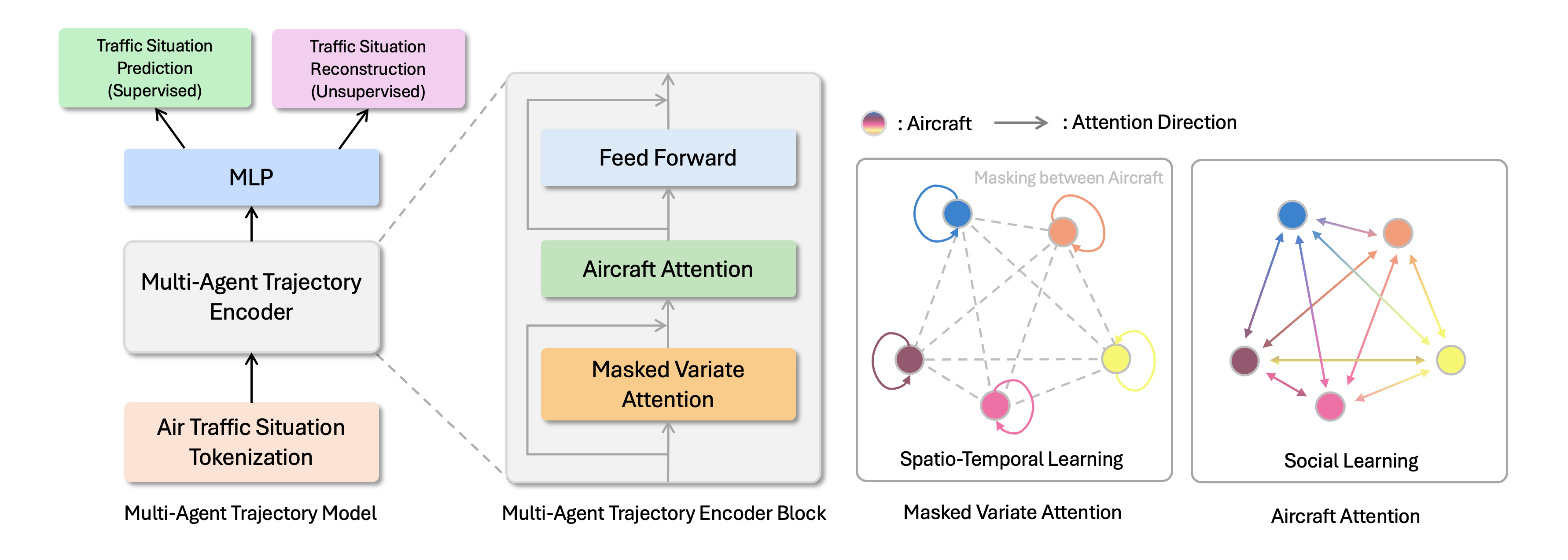}
    \vspace{-2mm} 
    \caption{\textbf{Overall architecture of the multi-agent trajectory model}. First, the air traffic situation is tokenized and prepared as input for the model. Next, the multi-agent trajectory encoder captures two key relationships in the air traffic situation: (i) spatio-temporal relationship (masked variate attention) and (ii) social relationship (aircraft attention). Finally, the encoded air traffic situation is decoded either for prediction (supervised) or reconstruction (unsupervised) through MLP.}

    \vspace{-5mm} 
    \label{frameworkfigure}
 \end{figure*}

\section{Proposed Framework}\label{framework}
To explain air traffic situations, a multi-agent trajectory model~\cite{yoon2025maiformer} is used that considers both individual aircraft motions as well as their interactions over time.  This model employs a Transformer-based architecture~\cite{vaswani2017attention} that includes a self-attention mechanism at its core to model relationships among aircraft. The comprehensive overview of the utilized multi-agent trajectory model is depicted in Figure~\ref{frameworkfigure}. 

In the model, we capture two relationships: (i) the spatio-temporal relationship of individual aircraft movements via masked variate attention, and (ii) the social relationship between multiple aircraft through aircraft attention. The first relationship sheds light on an aircraft’s evolution over time, while the second underscores how the movement of one aircraft influences others in the airspace. To explain air traffic situations, we primarily focus on the second relationship, as it yields insights into inter-aircraft interactions.

Let’s consider air traffic situation $X$ at timestep $t$, involving $N$ aircraft flying over a time horizon $T$, denoted as follows:
\begin{equation}
    X = \{\text{AC}_1,\text{AC}_2,\ldots, \text{AC}_N\}
\end{equation}
where $\text{AC}_i=[x_{t-T+1:t},y_{t-T+1:t},z_{t-T+1:t}]$ represents the series of the position vectors of each aircraft over time $T$. Once air traffic situation $X$ is fed into the model, the self-attention mechanism between aircraft can be applied as follows:
\begin{equation}
    \text{Attention}(Q,K,V)=\text{Softmax}(\frac{QK^{\top}}{\sqrt{d_k}})V
\end{equation}
where $Q,K$ and $V$ represent queries, keys and values, derived from the inputs $X$ through learnable transformations. Note that the entire trajectory sequence of each aircraft can be aggregated into a single token using inverted embedding~\cite{liu2023itransformer,Yoon2025Aircraft}. The dot product of $Q$ and $K$ then computes the degree of relationship between aircraft as numerical values.

We can use attention scores to interpret the influences between different aircraft and thus explain certain air traffic situations. When there are multiple aircraft in the airspace, attention scores are calculated for a selected aircraft relative to the others to evaluate how its behavior is being shaped by surrounding aircraft. This method enables us to identify aircraft with high attention scores, interpreting them as the most influential on the selected aircraft in that given situation.

We explore two learning paradigms to train our model: (i) the supervised setting, and (ii) the unsupervised setting. Under the supervised setting, the model is trained to use past air traffic scenes as input and predict future air traffic scenes. Meanwhile, in the unsupervised setting, the model is trained to take air traffic scenes as inputs and then reconstruct the same scenes as outputs. We selected the supervised setting for this work to ensure the model captures how future flight intentions are influenced by the surrounding traffic. For training and evaluation purposes, we relied upon a dataset consisting of four months of air traffic surveillance data from the terminal airspace surrounding Incheon International Airport (ICN), South Korea, gathered between January 2023 and April 2023.

\begin{figure*}[t!]			
	\centering
	\includegraphics[width=0.95\textwidth,height=0.95\textheight]{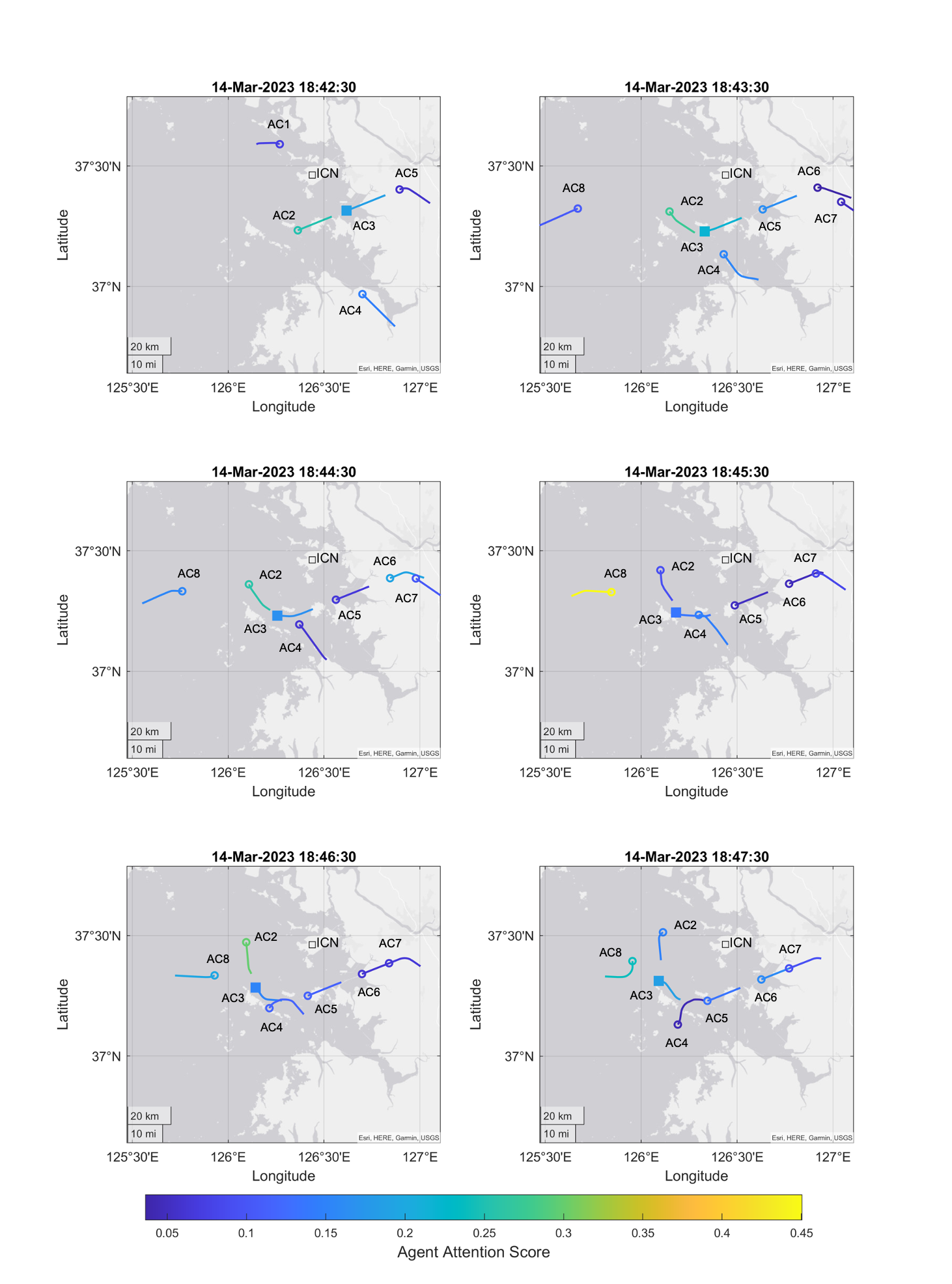}
    \caption{Illustrative explanation results for dynamic air traffic situation with attention scores.}
    \label{results}
 \end{figure*}

 \begin{figure*}[t!]			
	\centering
	\includegraphics[width=0.93\textwidth]{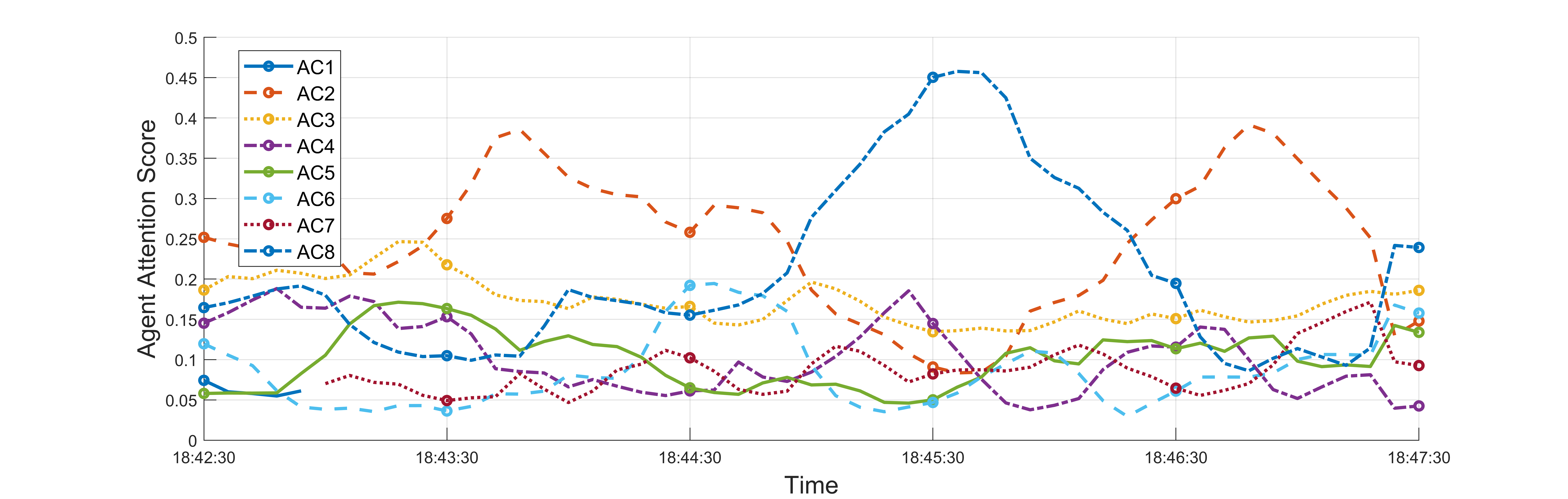}
    \caption{Attention scores of individual aircraft evolving over time in the above scenario in Figure~\ref{results}.}
    
    \vspace{-5mm} 
    \label{attentiontime}
 \end{figure*}

\section{Preliminary Results}\label{preliminary}
As an illustrative example of our work, we introduce a dynamic air traffic situation at ICN between 18:42:30 and 18:47:30 on March 14, 2023. Here, multiple aircraft are approaching runway 15L from three entry fixes (west, south, and east), as displayed in Figure~\ref{results}. Traffic flows from the south and east entry fixes of ICN’s terminal airspace typically merge early, forming structured patterns. In contrast, traffic flows from the west entry fix often bring complexity due to its proximity to ICN. This leads ATCo to frequently stretch or shortcut trajectories of aircraft from the west entry fix to maintain separation from arriving aircraft from other entry fixes. In the figure, each aircraft's current position is marked with a circle ({\raisebox{0.1ex}{\scalebox{0.8}{$\bigcirc$}}}), with its past two-minute trajectory appearing as a tail. The tail’s length reflects the aircraft’s horizontal speed. To highlight interactions, we color each trajectory based on attention scores: strong interactions are shown in yellow, while weaker interactions appear in blue.

This example centers on AC3, whose current position is marked with a square (\textcolor{black}{\raisebox{-0.2ex}{$\blacksquare$}}). It explains the air traffic situation from AC3’s viewpoint, primarily describing how other aircraft in the airspace influence AC3’s behavior. Attention scores are computed with AC3 as the query aircraft, while all other aircraft, including AC3 itself, serve as key aircraft. These scores are dynamically updated, offering insights into how interactions between aircraft evolve.

In the initial three timeframes (18:42:30 - 18:44:30), five aircraft are sequenced for arrival, forming a structured traffic flow without any apparent critical interactions. During these phases, AC3 primarily focuses on AC2, the preceding aircraft on the same Standard Terminal Arrival Route (STAR). Although AC8 enters the airspace from the opposite direction (west entry fix), AC3 does not appear to pay attention to AC8. AC8 seems to have a later landing order than AC3 during these time frames, suggesting it may not impact AC3’s subsequent behavior.

In the fourth frame (18:45:30), AC8 adjusts its heading to join the traffic flow from the east side. The ATCo provided AC8 with a direct-to-instruction, resulting in an earlier landing order for AC8, compared to AC3. Without this instruction, AC8 would need a significantly longer flight time to join the traffic flow. Consequently, AC3 pays significant attention to AC8, whose behavior greatly impacts AC3’s future trajectory.

As the landing sequence adjusts, AC3’s attention scores for other aircraft stabilize at a moderate or lower level. AC3 mainly focuses on itself and the preceding aircraft (AC2 and AC8) until the last frame (18:47:30). This suggests that no other aircraft influenced AC3’s behavior significantly during this period. In the final frame, AC4 diverges from the traffic flow, possibly to maintain the required separations from other incoming aircraft. Notably, our framework detects this change and AC3 assigns AC4 the lowest attention score, acknowledging that AC4 no longer impacts AC3’s future trajectory as their paths diverge. Figure~\ref{attentiontime} further demonstrates how the attention scores of individual aircraft evolve over time in the above scenario.

Through this example, we demonstrate the efficacy of our proposed machine learning-based framework in interpreting dynamic air traffic scenarios using attention scores. The framework allows us to identify alterations in traffic flow by visualizing changes in the aircraft’s attention scores. By spotlighting influential aircraft with high attention scores, our method provides insights into the ATCo’s decisions regarding a specific aircraft and its interactions with surrounding traffic. Although this paper primarily discusses the air traffic situation from the perspective of a single aircraft (Lagrangian perspective), our framework can also provide a view from a specific location in airspace (Eulerian perspective) by averaging the attention scores of all aircraft in the area.

\section{Discussion}\label{discussion}
We have demonstrated the potential of our approach in explaining air traffic situation using attention scores. While our model can effectively quantify the level of attention paid to aircraft in the airspace, the interpretability of these scores remains an open question. In particular, the relationship between machine attention and human ATCo attention is not yet fully understood. Nonetheless, prior works have explored the alignment between machine attention and human attention patterns~\cite{kozlova2024transformer,bensemann2022eye}, suggesting that our model’s attention mechanisms may reflect, at least in part, the cognitive patterns employed by ATCos. However, this hypothesis requires further empirical validation.

Even assuming such a correlation exists, the attention scores derived from our model represent only a partial and artificial approximation of the ATCo’s mental model. In practice, ATCo base their decision on a complex and multifaceted set of variables, including weather conditions, airspace structure, flight procedures, and other operational constraints—many of which are not currently modeled in the proposed framework. The next step of our research is to incorporate these additional factors into the framework and evaluate whether the enriched model better aligns with observed ATCo attention patterns, potentially via human-in-the-loop (HITL) experiments.

From an application standpoint, our framework may open pathways for both tactical and strategic applications within real-world air traffic operations. At the tactical level, by identifying aircraft that are likely to influence ATCo decision-making, the framework can deliver real-time support during high-density operations, mitigating cognitive workload by proactively highlighting aircraft that may require immediate attention. At the strategic level, patterns in attention scores accumulated over time can reveal hidden inefficiencies in airspace design or traffic flow dynamics. For instance, if certain regions or routes consistently attract disproportionate attention, such trends may indicate imbalances in workload distribution. These insights can guide procedural revisions or airspace redesigns to promote more equitable and efficient ATCo workload allocation.

\section{Conclusion and Future Works}\label{conclusion}
In this study, we propose a machine learning-based framework to elucidate dynamic air traffic situations. Our approach employs a multi-agent trajectory model that captures both individual aircraft behavior and interaction patterns between aircraft. We have validated our framework using actual air traffic surveillance data from the terminal airspace surrounding Incheon International Airport in South Korea. Our framework explains air traffic situations by quantifying the influences of one aircraft on others through the model’s attention scores. The results indicate that our method effectively deciphers air traffic situations by emphasizing the aircraft that are most influential in the scenarios.

In our future work, we aim to go beyond explaining air traffic scenarios, striving to achieve a deeper comprehension of these situations. This will enable us to uncover the complex and abstract decision-making patterns of air traffic controllers. Such insights could not only facilitate the inference of flight intentions but also those of the controllers, all the while offering plausible actions that align with their historical control behaviors. Moreover, to validate the generalizability of our approach across diverse operational contexts, future work should explore its application to General Aviation (GA) and Visual Flight Rules (VFR) operations.

\bibliographystyle{IEEEtran}
\bibliography{mybib.bib}

@article{kuchar2000review,
  title={A review of conflict detection and resolution modeling methods},
  author={Kuchar, James K and Yang, Lee C},
  journal={IEEE Transactions on intelligent transportation systems},
  volume={1},
  number={4},
  pages={179--189},
  year={2000},
  publisher={IEEE}
}

@article{Yoon2025Aircraft,
  title={Aircraft Trajectory Prediction with Inverted Transformer},
  author={Yoon, Seokbin and Lee, Keumjin},
  journal={IEEE Access},
  year={2025},
  publisher={IEEE}
}

@inproceedings{liu2023itransformer,
  title={iTransformer: Inverted Transformers Are Effective for Time Series Forecasting},
  author={Liu, Yong and Hu, Tengge and Zhang, Haoran and Wu, Haixu and Wang, Shiyu and Ma, Lintao and Long, Mingsheng},
  booktitle={International Conference on Learning Representations},
year={2024}
}

@article{pallottino2002conflict,
  title={Conflict resolution problems for air traffic management systems solved with mixed integer programming},
  author={Pallottino, Lucia and Feron, Eric M and Bicchi, Antonio},
  journal={IEEE transactions on intelligent transportation systems},
  volume={3},
  number={1},
  pages={3--11},
  year={2002},
  publisher={IEEE}
}

@article{beasley2000scheduling,
  title={Scheduling aircraft landings—the static case},
  author={Beasley, John E and Krishnamoorthy, Mohan and Sharaiha, Yazid M and Abramson, David},
  journal={Transportation science},
  volume={34},
  number={2},
  pages={180--197},
  year={2000},
  publisher={INFORMS}
}

@article{histon2002introducing,
  title={Introducing structural considerations into complexity metrics},
  author={Histon, Jonathan M and Hansman, R John and Aigoin, Guillaume and Delahaye, Daniel and Puechmorel, Stephane},
  journal={Air Traffic Control Quarterly},
  volume={10},
  number={2},
  pages={115--130},
  year={2002},
  publisher={American Institute of Aeronautics and Astronautics, Inc.}
}

@inproceedings{bilimoria2005analysis,
  title={Analysis of aircraft clusters to measure sector-independent airspace congestion},
  author={Bilimoria, Karl and Lee, Hilda},
  booktitle={AIAA 5th ATIO and16th Lighter-Than-Air Sys Tech. and Balloon Systems Conferences},
  pages={7455},
  year={2005}
}

@article{porretta2008performance,
  title={Performance evaluation of a novel 4D trajectory prediction model for civil aircraft},
  author={Porretta, Marco and Dupuy, Marie-Dominique and Schuster, Wolfgang and Majumdar, Arnab and Ochieng, Washington},
  journal={The Journal of Navigation},
  volume={61},
  number={3},
  pages={393--420},
  year={2008},
  publisher={Cambridge University Press}
}

@inproceedings{hong2015modeling,
  title={Modeling the air traffic controller’s direct-to operation using logistic regression},
  author={Hong, Sungkweon and Jung, Soyeon and Lee, Keumjin},
  booktitle={15th AIAA Aviation Technology, Integration, and Operations Conference},
  pages={2906},
  year={2015}
}

@article{guleria2024towards,
  title={Towards conformal automation in air traffic control: Learning conflict resolution strategies through behavior cloning},
  author={Guleria, Yash and Pham, Duc-Thinh and Alam, Sameer and Tran, Phu N and Durand, Nicolas},
  journal={Advanced Engineering Informatics},
  volume={59},
  pages={102273},
  year={2024},
  publisher={Elsevier}
}

@article{jung2018data,
  title={A data-driven air traffic sequencing model based on pairwise preference learning},
  author={Jung, Soyeon and Hong, Sungkweon and Lee, Keumjin},
  journal={IEEE Transactions on Intelligent Transportation Systems},
  volume={20},
  number={3},
  pages={803--816},
  year={2018},
  publisher={IEEE}
}

@article{vaswani2017attention,
  title={Attention is All You Need},
  author={Vaswani, Ashish and Shazeer, Noam and Parmar, Niki and Uszkoreit, Jakob and Jones, Llion and Gomez, Aidan N and Kaiser, {\L}ukasz and Polosukhin, Illia},
  journal={Advances in Neural Information Processing Systems},
  volume={30},
  year={2017}
}

@article{yoon2025maiformer,
  title={Multi-Agent Inverted Transformer for Flight Trajectory Prediction},
  author={Yoon, Seokbin and Lee, Keumjin},
  journal={IEEE Transactions on Intelligent Transportation Systems},
  year={2026},
  publisher={IEEE}
}

@inproceedings{nunes2003identifying,
  title={Identifying controller strategies that support the ‘Picture’},
  author={Nunes, Ashley and Mogford, Richard H},
  booktitle={Proceedings of the Human Factors and Ergonomics Society Annual Meeting},
  pages={71-75},
  year={2003}
}

@inproceedings{kozlova2024transformer,
  title={Transformer Attention vs Human Attention in Anaphora Resolution},
  author={Kozlova, Anastasia and Akhmetgareeva, Albina and Khanova, Aigul and Kudriavtsev, Semen and Fenogenova, Alena},
  booktitle={Proceedings of the Workshop on Cognitive Modeling and Computational Linguistics},
  pages={109--122},
  year={2024}
}

@inproceedings{bensemann2022eye,
  title={Eye gaze and self-attention: How humans and transformers attend words in sentences},
  author={Bensemann, Joshua and Peng, Alex and Benavides-Prado, Diana and Chen, Yang and Tan, Neset and Corballis, Paul Michael and Riddle, Patricia and Witbrock, Michael J},
  booktitle={Proceedings of the Workshop on Cognitive Modeling and Computational Linguistics},
  pages={75--87},
  year={2022}
}
\vspace{12pt}

\end{document}